# CAR-Net: Clairvoyant Attentive Recurrent Network

Amir Sadeghian[1], Ferdinand Legros[1*], Maxime Voisin[1*], Ricky Vesel[2],
Alexandre Alahi[3], Silvio Savarese[1]

[1]Stanford University, [2] Race Optimal,
[3]Ecole Polytechnique Federale de Lausanne (EPFL), Switzerland
{amirabs,flegros,maxime.voisin,ssilvio}@stanford.edu;
vesel.rw@gmail.com; alexandre.alahi@epfl.ch

**Abstract.** We present an interpretable framework for path prediction that leverages dependencies between agents' behaviors and their spatial navigation environment. We exploit two sources of information: the past motion trajectory of the agent of interest and a wide top-view image of the navigation scene. We propose a Clairvoyant Attentive Recurrent Network (CAR-Net) that learns where to look in a large image of the scene when solving the path prediction task. Our method can attend to any area, or combination of areas, within the raw image (*e.g.*, road intersections) when predicting the trajectory of the agent. This allows us to visualize fine-grained semantic elements of navigation scenes that influence the prediction of trajectories. To study the impact of space on agents' trajectories, we build a new dataset made of top-view images of hundreds of scenes (Formula One racing tracks) where agents' behaviors are heavily influenced by known areas in the images (*e.g.*, upcoming turns). CAR-Net successfully attends to these salient regions. Additionally, CAR-Net reaches state-of-the-art accuracy on the standard trajectory forecasting benchmark, Stanford Drone Dataset (SDD). Finally, we show CAR-Net's ability to generalize to unseen scenes.

## 1 Introduction

Path prediction consists in predicting the future positions of agents (*e.g.*, humans or vehicles) within an environment. It applies to a wide range of domains from autonomous driving vehicles [1] and social robot navigation [2–4], to abnormal behavior detection in surveillance [5–10]. Observable cues relevant to path prediction can be grouped into dynamic and static information. The former captures the previous motion of all agents within the scene (past trajectories). The latter consists of the static scene surrounding agents [11–13]. In this work, we want to leverage the static scene context to perform path prediction. The task is formulated as follows: given the past trajectory of an agent (x-y coordinates of past few seconds) and a large visual image of the environment (top-view of the scene), we want to forecast the trajectory of the agent over the next few seconds. Our model should learn where to look within a large visual input to enhance its prediction performance (see Fig. 1).

Predicting agents' trajectories while taking into account the static scene context is a challenging problem. It requires understanding complex interactions between agents and space, and encoding these interactions into the path prediction model. Moreover, scene-specific cues are often sparse and small within the visual input, *e.g.*, a traffic sign within the scene. Finally, these cues might be far from the agent of interest.

---

* indicates equal contribution

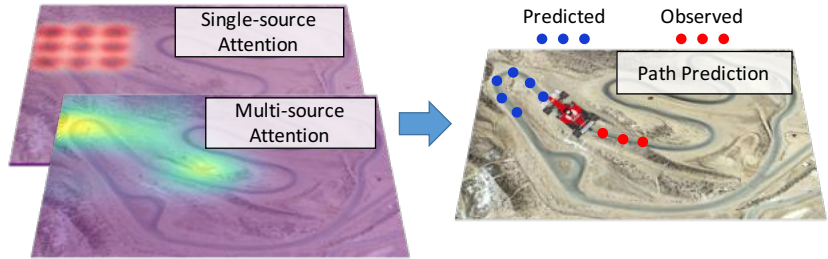

**Fig. 1. CAR-Net** is a deep attention-based model that combines two attention mechanisms for path prediction.

Recent research in computer vision has successfully addressed some of the challenges in path prediction. Kitani et al. [14] have demonstrated that the semantic segmentation of the environment (*e.g.*, location of sidewalks and grass areas) helps to predict pedestrian trajectories. Ballan *et al.* [15] modeled human-space interactions using navigation maps that encode previously observed scene-specific motion patterns. These methods rely on scene semantic information collected in advance. Instead, our method relies on raw images, which are easier to obtain, and our method has the potential to infer finer-grained scene semantics and functional properties of the environment. To this end, Lee *et al.* [16] have used raw images to predict agents' trajectories. However, their method does not provide a way to understand what visual information within the scene is "used" by the model to predict future trajectories.

We address the limitations of previous path prediction methods by proposing a visual attention model that leverages agent-space interactions and enhances prediction accuracy. Inspired by the recent use of attention models and neural networks in image captioning [17], machine translation [18], knowledge bases [19, 20], and object recognition [21, 22], we introduce the first visual attention model that can predict the future trajectory of an agent while attending to the salient regions of the scene. Our method is able to attend to any region, or collection of regions, in the image. Attention based models can be broadly categorized into single and multi-source attention models. Single source attention models (*e.g.*, DRAW [23, 21]) attend to features extracted from a single area of the image, while multi-source attention models (*e.g.*, soft attention from [17]) use a combination of features from all areas of the image. In this paper, we propose CAR-Net, a deep neural network architecture which predicts future trajectories - hence being *Clairvoyant* - by processing raw top-view images with a visual *Attentive Recurrent* component. Our attention model combines both single-source and multi-source attention mechanisms. By leveraging both attention mechanisms, our prediction framework makes use of a wider spectrum of agent-space dependencies. Moreover, CAR-Net is simple to implement and train. Hence, it facilitates the use of trajectory prediction in a wide range of other vision tasks such as object tracking [5], activity forecasting [24] and action localization [25].

To study if our proposed architecture is able to learn observable agent-space correlations, we build a new dataset where agents' behaviors are largely influenced by known regions within a scene (*e.g.*, a curve in the road). As opposed to other popular datasets for trajectory prediction, the proposed dataset allows to understand the effect of the environment on agents' future trajectories. Because the dataset is composed of static scenes, future trajectories are not affected by confounding factors such as the behavior of other agents. This disentangles the contributions of scene semantic information and of other agents' interactions, in the task of path prediction. To build this new dataset, we have collected more than two hundred real-world Formula One racing tracks and computed the vehicles' optimal paths given the tracks' curvatures using equations in [26]. In this context, the geometry of the road causes the vehicle to speed up or down, and steer. Our attention mechanism succeeds at leveraging elements of the tracks, and effectively predicts the optimal paths of vehicles on these tracks. As part of our contributions, this new dataset for path prediction and learning agent-space correlations will be released publicly. We further show that the accuracy of our method outperforms previous approaches on the Stanford Drone Dataset (SDD), a publicly available trajectory forecasting benchmark where multiple classes of agents (*e.g.*, humans, bicyclists, or buses) navigate outdoor scenes. CAR-Net is an intuitive and simple model that achieves state-of-the-art results for path prediction, while enabling the visualization of semantic elements that influenced prediction thanks to attention mechanisms.

## 2 Related Work

**Trajectory forecasting.** Path prediction given the dynamic content of a scene has been extensively studied with approaches such as Kalman filters [27], linear regressions [28], or non-linear Gaussian Processes [29–31, 2]. Pioneering work from Helbing and Molnar [32–34] presented a pedestrian motion model with attractive and repulsive forces referred to as the Social Force model. All these prior works have difficulty in modeling complex interactions. Following the recent success of Recurrent Neural Networks (RNN) for sequence prediction tasks, Alahi *et al.* [35, 36] proposed a model which learns human movement from data to predict future trajectories. Recently, Robicquet *et al.* [37, 38] proposed the concept of social sensitivity with a social force based model to improve path forecasting. Such models suffice for scenarios with few agent-agent interactions, but they do not consider agent-space interactions. In contrast, our method can handle more complex environments where agents' behaviors are severely influenced by scene context (*e.g.*, drivable road vs trees and grass).

Recent works have studied how to effectively leverage static scenes in the path prediction task. Kitani *et al.* [14] used semantic knowledge of the scene to forecast plausible paths for a pedestrian using inverse optimal control (IOC). Walker *et al.* [1] predicted the behavior of generic agents (*e.g.*, vehicles) in a scene given a large collection of videos, but in a limited number of scenarios. Ballan *et al.* [15] learned scene-specific motion patterns and applied them to novel scenes with an image-based similarity function. Unfortunately, none of these methods can provide predictions using raw images of scenes. Recently, Lee *et al.* [16] proposed a method for path prediction given the scene context using raw images. However, all these methods have limited interpretability. Our method is instead designed for this specific purpose: providing an intuition as to why certain paths are predicted given the context of the scene.

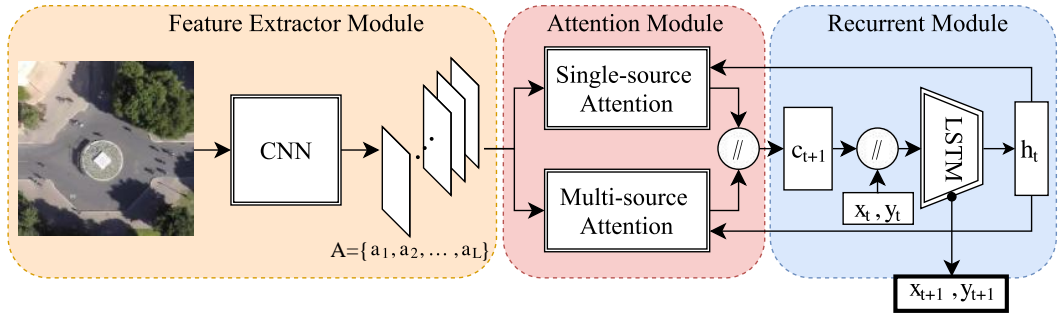

**Fig. 2.** Overview of CAR-Net architecture. Note that "//" block is concatenation operation.

**Visual Attention.** Related work from Xu and Gregor [17, 23] introduces attention based models that learn to attend the salient objects related to the task of interest. Xu et al. [17], present soft and hard attention mechanisms that attend to the entire image. Soft attention applies a mask of weights to the image's feature maps. Since the associated training operation is differentiable, it has been applied to a wide range of tasks. The hard attention mechanism is not differentiable and it must be trained by Reinforcement Learning. The non-differentiability of this method has led to scarce applications.

Other attention models apply dimensionality reduction to the image. Their goal is to accumulate information over a sequence of partial glimpses of the image. The recurrent attention model introduced in [21] attends to a sequence of crops of the image. It has been used in many tasks such as digit classification and person identification [39, 23, 40]. Visual attention models have also been widely applied to many other applications such as image classification [41], image captioning [17, 42], and video classification [43]. Inspired by these works, we hereby use visual attention mechanisms in our model to perform trajectory prediction.

## 3  CAR-Net

Scene context is necessary to predict the future behavior of agents. For instance, a cyclist approaching a roundabout changes his path to avoid collision. Such deviations in trajectories cannot be predicted by only observing the agent's past positions. This motivates us to build a model that can leverage observable scene context while predicting an agent's future path. We introduce CAR-Net, a deep attention-based model for path prediction. It performs trajectory prediction using raw top-view images of scenes and past trajectories of agents. CAR-Net is able to attend to the most relevant parts of the input image. In this section we first describe the overall architecture of our model. Then, we explain our visual attention module.

### 3.1  Overall Architecture

The objective of our model is to predict the future path of an agent given its past trajectory and a top-view image of the scene. Our model uses a *feature extractor* to derive feature vectors from the raw image (Fig. 2). Then, a *visual attention module* computes a context vector $c_t$ representing the salient areas of the image to attend at time $t$. Finally, in the *recurrent module*, a long short-term memory (LSTM) network [44] generates the future position of the agent $(x_{t+1}, y_{t+1})$ at every time step, conditioned on the context vector $c_t$, on the previous hidden state $h_t$, and on the previously generated position of the agent $(x_t, y_t)$. Our model is able to capture agent-space interactions by combining both the scene context vector and the past trajectory of the agent.

### 3.2   Feature extractor module

We extract feature maps from static top-view images using a Convolutional Neural Network (CNN). We use VGGnet-19 [45] pre-trained on ImageNet [46] and fine-tuned on the task of scene segmentation as described in [47]. Fine-tuning VGG on scene segmentation enables the CNN to extract image features that can identify obstacles, roads, sidewalks, and other scene semantics that are essential for trajectory prediction. We use the output of the $5^{th}$ convolutional layer as image features. The CNN outputs $L = N \times N$ feature vectors, $A = \{a_1, ..., a_L\}$, of dimension D, where N and D are the size and the number of feature maps outputted by the $5^{th}$ convolutional layer, respectively. Each feature vector corresponds to a certain region of the image. Fig. 2 depicts the feature extractor module.

### 3.3   Visual attention module

Given a high-dimensional input image of a scene, we want our model to focus on smaller, discriminative regions of this input image. Using a visual attention method, the most relevant areas of the image are extracted while irrelevant parts are ignored. The general attention process works as follows. A layer $\phi$ within the attention mechanism takes as input the previous hidden state $h_t$ of the LSTM and outputs a vector $\phi(h_t)$ that is used by the attention mechanism to predict the important areas of the image. The vector $\phi(h_t)$ is then applied to feature vectors $A$ (through a function $f_{att}$), resulting in a context vector $c_{t+1}$ that contains the salient image features at time step $t + 1$:

$$c_{t+1} = f_{att}(A, \phi(h_t)). \qquad (1)$$

Our visual attention module can be substituted with any differentiable attention mechanism. Moreover, it can use a combination of several attention methods. Provided that $f_{att}$ and $\phi$ are differentiable, the whole architecture is trainable by standard back-propagation. We propose three variants for the differentiable attention module that are easily trainable. The first method extracts visual information from multiple areas of the image with a soft attention mechanism. The second method extracts local visual information from a single cropped area of the image with an attention mechanism inspired by [23]. We refer to the first and second methods as *multi-source* and *single-source* attention mechanisms, respectively. Finally, the attention module of CAR-Net combines both attention mechanisms, allowing our prediction framework to learn a wider spectrum of scene dependencies.

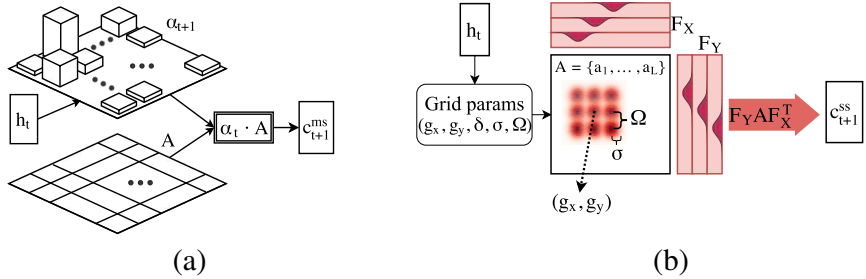

**Fig. 3.** Our multi and single-source attention mechanism

**CAR-Net attention.** Learning agent-space interactions, and encoding them into the path prediction model is a challenging task. The scene-specific cues are sometimes sparse and spread throughout the entire image far away from the agent, or small within a specific area of the image. Single and multi-source attention mechanisms attend respectively to localized and scattered visual cues in the scene. When the relevant visual cues are scattered all over the input image, a multi-source attention method can successfully extract a combination of features from multiple key areas of the image. In contrast, when the relevant visual information is localized in one particular area of the image, single-source attention methods are a good fit to attend to that specific region. Note that multi-source attention does not necessarily reduce to single-source attention and they complement each other.

To leverage both local and scattered visual cues in path prediction, the core attention module in CAR-Net combines the two context vectors obtained from single and multi-source attention mechanisms. The combination is done by concatenating the context vectors from single-source $c_t^{ss}$ and multi-source $c_t^{ms}$ attention mechanisms, $c_t = [c_t^{ss}, c_t^{ms}]$. The attention module in Fig. 2 depicts the process. More technical details about multi and single-source attention mechanisms can be found in Sec. 3.3. CAR-Net outperforms both single and multi-source attention mechanisms, proving its ability to leverage the strengths of the two attention mechanisms.

**Multi-source attention.** The multi-source attention mechanism applies weights to all spatial areas of the scene based on their importance, and outputs a context vector containing relevant scene context from multiple regions of the image. First, the weights matrix $\alpha_{t+1}$ is calculated by passing the hidden state $h_t$ through a fully connected layer $\phi$ with weight $W_{ms}$ and bias $b_{ms}$. Later, the context vector $c_{t+1}^{ms}$ is calculated by element-wise product of the weight matrix $\alpha_{t+1}$ and the feature maps $A$. Fig. 3(a) and Eq. 2 show the entire process:

$$
\begin{aligned}
c_{t+1}^{ms} &= f_{att}(A, \phi(h_t)) \\
&= a \cdot \phi(h_t) = A \cdot \alpha_{t+1} \\
\alpha_{t+1} &= softmax(W_{ms}h_t + b_{ms}).
\end{aligned}
\tag{2}
$$

The soft (multi-source) attention mechanism described in [17] calculates the weight matrix $\alpha_{t+1}$ conditioned on both the previous hidden vector and the features of the image. However, our $\alpha_{t+1}$ relies only on the previous hidden vector. The distinction is important because for path prediction tasks, we do not have the future images of a scene. Moreover, it reduces the computation cost without impacting the performance of the model.

**Single-source attention.** The single-source attention mechanism illustrated in Fig. 3(b) attends to a single local area in the image. To do so, we adapt the DRAW attention mechanism - which was initially designed for the unsupervised setting of digit generation [23] - to the supervised learning setting of path prediction. The single-source attention mechanism attends to the region of the image defined by a local grid of $N$ Gaussians. The center $(g_X, g_Y)$ of the grid, the stride $\delta$ of the grid, and the variance $\sigma$ of all $N$ Gaussians, are predicted by the model at each time step $t + 1$ by mapping linearly the hidden state $h_t$ to attention parameters $(g_X, g_Y, \delta, \sigma)$. The stride of the grid controls the "zoom" of the local area attended by the model. As the stride gets larger, the grid of Gaussians covers a larger area of the original image. The exact position $(\nu_x^i, \nu_y^i)$ of each Gaussian $i$ on the grid is found using the center and the stride of the grid as in Eq. 3.

$$
\begin{aligned}
\nu_X^i &= g_X + (i - N/2 - 0.5)\delta \\
\nu_Y^i &= g_Y + (i - N/2 - 0.5)\delta
\end{aligned}
\tag{3}
$$

The resulting grid of Gaussians defines two filter-bank matrices $F_X$ and $F_Y$, using Eq. 4. Using these filter-bank matrices, the single-source attention mechanism is able to attend the region of the image defined by the local grid of Gaussian: $F_X$ and $F_Y$ are convoluted with the feature maps $A$ of the image, as in Eq. 5. The resulting context vector $c_{t+1}^{ss}$ contains scene context from to the single local area of the image corresponding to the grid of Gaussians.

$$
\begin{aligned}
F_X[i, a] &= \frac{1}{Z_X} exp\Big( -\frac{(a - \nu_X^i)^2}{2\sigma^2} \Big) \\
F_Y[j, b] &= \frac{1}{Z_Y} exp\Big( -\frac{(b - \nu_Y^j)^2}{2\sigma^2} \Big).
\end{aligned}
\tag{4}
$$

$$
c_{t+1}^{ss} = f_{att}(A, \phi(h_t)) = F_X(h_t)^T A F_Y(h_t).
\tag{5}
$$

Note that indexes $(i, j)$ refer to Gaussians in the grid and that indexes $(a, b)$ refer to locations in the feature maps. The normalization constants $Z_x$, $Z_y$ ensure $\sum_a F_X[i, a] = 1$ and $\sum_b F_Y[j, b] = 1$.

### 3.4   Implementation details

We trained the LSTM and the attention module from scratch with the Adam optimizer [48], a mini-batch size of 128, and a learning rate of 0.001 sequentially decreased every 10 epochs by a factor of 10. All models are trained for 100 epochs, on the L2 distance between ground-truth and predicted trajectories. As in many sequence prediction tasks, the training and testing process is slightly different. At training time, the ground-truth positions are fed as inputs to the LSTM. In contrast, at test time, the predictions of positions $(x_t, y_t)$ are re-injected as inputs to the LSTM at the next time step.

## 4   Experiments

We presented CAR-Net, a framework that provides accurate path predictions by leveraging spatial scene contexts. We perform a thorough comparison of our method to state-of-the-art techniques along with comprehensive ablation experiments. We then present insights on the interpretability of our method. We finally show the generality and robustness of CAR-Net by experimenting with different datasets.

### 4.1   Data

We tested our models on the following three datasets that all include trajectory data and top-view images of navigation scenes.

**Stanford Drone Dataset (SDD) [37].** To show that CAR-Net achieves state-of-the-art performance on path prediction, we tested the model on SDD, a standard benchmark for path prediction [16, 35, 37]. This large-scale dataset consists of top-view videos of various targets (*e.g.*, pedestrians, bicyclists, cars) navigating in many real-world outdoor environments in a university campus (20 different scenes). Trajectories were split into segments of 20 time steps each (8s total), yielding approximately 230K trajectory segments. Each segment is composed of 8 past positions (3.2s), which are fed to the network as sequential inputs, and 12 future positions (4.8s) used to evaluate predictions. This is the standard temporal setup for path prediction on SDD. We use raw images to extract visual features, without any prior semantic labeling. We adopt the standard benchmark dataset split for SDD.

**Formula One Dataset.** Studying the influence of space on agents' trajectories is complex since agents' behaviors are influenced not only by the semantics of the navigation scene, but also by other factors such as interactions with other agents. For instance, a pedestrian could stop as they meet an acquaintance. We release a Formula One (F1) dataset, composed of real-world car racing tracks and their associated trajectories. This dataset provides a controlled environment to evaluate how well models can extract useful spatial information for trajectory prediction. In the F1 dataset, the agents' behaviors can be largely explained by the geometry of the tracks (*e.g.* the curve of an upcoming turn). As opposed to other popular datasets for trajectory prediction (*e.g.* SDD), F1 dataset allows for evaluations in static settings where future trajectories are not affected by confounding factors such as the behavior of other agents. This disentangles the contributions of the spatial information and of other agents interactions in the task of trajectory prediction.

The top-view racing track images were obtained from Google Maps. On top of them, we simulated trajectories corresponding to an optimal driving pattern, referred to as "optimal trajectories" and computed with the equations presented in [26]. We used hand-segmented roads as inputs for the computations of optimal trajectories. Note that those optimal trajectories illustrate complex navigational patterns that depend on far away scene dependencies. The F1 dataset includes 250 tracks and more than 100K trajectories from different cities of Brazil, Canada, Columbia, Mexico, France, USA and other countries and will be available to the public for research purposes. Sample tracks are shown in Fig. 4. Car trajectories are split into 24 time step segments: 8 input past positions and 16 future positions used for evaluation. We opted for 16 future positions for evaluation, rather than 12 like in SDD, since the prediction task is simpler due to stronger agent-space dependencies. We split racing tracks in the F1 dataset into an 80% train set, 10% validation set, and 10% test set. The test racing tracks are totally unseen locations, they do not overlap with the training nor validation set racing tracks.

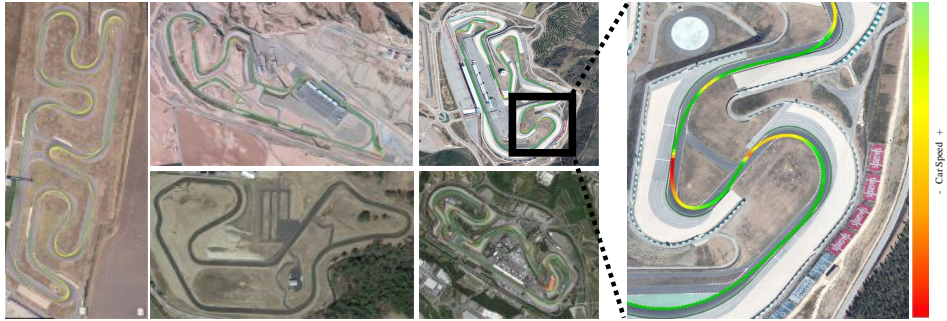

**Fig. 4.** Examples of scenes captured in the proposed F1 dataset. We annotated each track with the associated optimal racing trajectory.

**Car-Racing Dataset [49].** In order to derive further insights on how agent-space dependencies influence our model's predictions, we experimented with the Car-Racing dataset, a simpler racing track dataset that we synthesized. The Car-Racing dataset is composed of 3,000 tracks of various curvatures and road widths that we generated with the Car-Racing-v0 simulator from the OpenAI gym. We simulated (1) the optimal trajectories for each circuits, and (2) trajectories following the middle of the road at constant speed. Racing trajectories were split into 24 time-step segments, 8 input past positions and 16 future positions used for evaluation, yielding approximately 500K segments. We split racing tracks in this dataset into an 80% train set, 10% validation set, and 10% test set, which do not overlap.

**Optimal racing trajectories.** The ideal racing trajectory used in Car-Racing and F1 dataset is defined as the trajectory around a track that allows a given vehicle to traverse the track in minimum time. To calculate these optimal race trajectories, we segmented the roads by hand and computed the associated optimal racing paths using physics simulation. These simulations are based on 2D physical models from [26, 50].

## 4.2   Evaluation Metrics and Baselines

We measure the performance of our models on the path prediction task using the following metrics: (i) average displacement error - the mean L2 distance (ML2) over all predicted points of the predicted trajectory and the ground-truth points, (ii) final L2 distance error (FL2) - the L2 distance between the final predicted position and the final ground-truth position.

To perform an ablation study in Section 4.3 and show that our model achieves state-of-the-art performance in Section 4.4, we compare CAR-Net to the following baselines and previous methods from literature:

| Model | Car-Racing Middle | | Car-Racing Optimal | | Formula 1 Optimal | |
|---|---|---|---|---|---|---|
| | ML2 | FL2 | ML2 | FL2 | ML2 | FL2 |
| *T-LSTM* | 10.4 | 15.5 | 5.84 | 10.2 | 21.2 | 41.3 |
| *I-LSTM* | 9.71 | 14.1 | 5.62 | 9.5 | 20.8 | 40.1 |
| *MS-LSTM* | 7.35 | 12.7 | 5.30 | 8.71 | 18.9 | 37.8 |
| *SS-LSTM* | 6.36 | 9.91 | 4.64 | 7.63 | 14.7 | 28.9 |
| *CAR-Net* | **5.0** | **8.87** | **3.58** | **6.79** | **13.3** | **25.8** |

Table 1: Quantitative results of our methods on the Car-Racing dataset with middle and optimal trajectories and the F1 dataset. We report the Mean L2 error (ML2) and the Final L2 error (FL2). CAR-Net outperforms all models by combining single-source and multi-source attention outputs.

- *Linear model* (**Lin.**) We use an off-the-shelf linear predictor to extrapolate trajectories under the assumption of linear velocity.
- *Social Forces* (**SF**) and *Social-LSTM* (**S-LSTM**). We use the implementation of the Social Forces model from [51] where several factors such as group affinity have been modeled. Since the code for Social-LSTM is not available we compare our models with a self-implemented version of Social-LSTM from [35].
- *Trajectory only LSTM* (**T-LSTM**) and *Whole Image LSTM* (**I-LSTM**). These models are simplified versions of our model where we remove the image information and attention module, respectively.
- *Multi-Source only LSTM* (**MS-LSTM**) and *Single-Source only LSTM* (**SS-LSTM**). Our models using only multi-source attention and single-source attention mechanisms, respectively.
- *DESIRE*. A deep IOC framework model from [16]. We report the performance of the model *DESIRE-SI-IT0 Best* with top 1 sample.

### 4.3    Ablation Study

We performed an ablation study to show that prediction accuracy improves when combining single-source and multi-source attention mechanisms, which suggests that they extract complementary semantic cues from raw images. We analyzed the performances of baseline models and of CAR-Net on the racing track datasets (Car-Racing and Formula One datasets). We present our results in Table 1.

We observe similar results on both racing track datasets. First, I-LSTM only slightly outperforms T-LSTM. This seems to be because the large feature maps extracted from each racing track are too complex to significantly complement the dynamic cues extracted from agents' past trajectories. Second, attention models (MS-LSTM, SS-LSTM, CAR-Net) greatly outperform I-LSTM. This suggests that visual attention mechanisms enhance performance by attending to specific areas of the navigation scenes. We show in Section 4.5 that these attended areas are relevant semantic elements of navigation scenes - *e.g.* an upcoming turn. Note that SS-LSTM achieves lower errors than MS-LSTM. This is due to racing track images being large, and relevant semantic cues being mostly located close to the car. Finally, CAR-Net outperforms both MS-LSTM and SS-LSTM on all datasets. We think it is due to robustly combining the outputs of single-source and multi-source attention mechanisms.

| Model | ML2 | FL2 |
|---|---|---|
| *Lin.* | 37.11 | 63.51 |
| *SF [51]* | 36.48 | 58.14 |
| *DESIRE-SI-IT0 Best [16]* | 35.73 | 63.35 |
| *S-LSTM [35]* | 31.19 | 56.97 |
| *T-LSTM* | 31.96 | 55.27 |
| *I-LSTM* | 30.81 | 54.21 |
| *MS-LSTM* | 27.38 | 52.69 |
| *SS-LSTM* | 29.20 | 63.27 |
| *CAR-Net* | **25.72** | **51.80** |

Table 2: Performance of different baselines on predicting 12 future positions from 8 past positions on SDD. We report the Mean L2 error (ML2) and the Final L2 error (FL2) in pixel space of the original image. Our method, CAR-Net, achieves by far the lowest error.

**General remarks**. For the Car-Racing dataset, models perform better on the prediction of optimal trajectories than middle trajectories. This is due to the average pixel distance between consecutive positions being larger for middle trajectories than for the optimal trajectories. Also, we trained the models on 1K tracks for middle trajectories, instead of 3K for optimal trajectories.

### 4.4    Trajectory Forecasting Benchmark

CAR-Net outperforms state-of-the-art methods on the task of predicting 12 future positions (4.8s of motion) from 8 past positions (3.2s) on SDD benchmark, as reported in Table 2 (both lower ML2 and FL2 error). Note that the performance of *DESIRE-SI-IT0 Best* in [16] is provided for the task of predicting 4s of motion. We linearly interpolated this performance to obtain its performance on predicting 4.8s of motion and reported the interpolated number in Table 2.

The T-LSTM baseline achieves a lower ML2 error than Linear, SF, and S-LSTM models. However, the gaps between the FL2 errors of T-LSTM and SF or S-LSTM models are narrow, suggesting that the T-LSTM model tends to be relatively inaccurate when predicting the last future time-steps. We observe that S-LSTM performs poorly compared to MS-LSTM - especially in terms of FL2 error. We believe multi-source attention performs better due to scattered key semantics in SDD scenes. In all experiments, CAR-Net outperforms baselines methods regarding all metrics. Moreover, our model outperforms DESIRE with top 1 sample (DESIRE Best). This is consistent with [16] suggesting that regression-based models such as CAR-Net are a better fit for use cases where regression accuracy matters more than generating a probabilistic output.

**Generalization to unseen locations.**  CAR-Net generalizes to unseen locations in all datasets. This suggests that our model leverages observable scene features, rather than location-specific information. First, CAR-Net achieves better accuracy than other baseline methods on F1 test set, which is exclusively composed of unseen F1 racing tracks. Second, 9/17 (53%) locations in the SDD test set are unseen. The remaining 8/17 (47%) locations in the SDD test set are visually similar to training locations (*seen* locations). We evaluate our trained model's performance on the seen and on the unseen SDD test locations, separately. CAR-Net achieves similar performances on both *seen* and *unseen* test SDD locations - mean L2 distance of 23.87 and 26.93 pixels on the seen and unseen locations, respectively - proving its ability to generalize to unseen SDD locations.

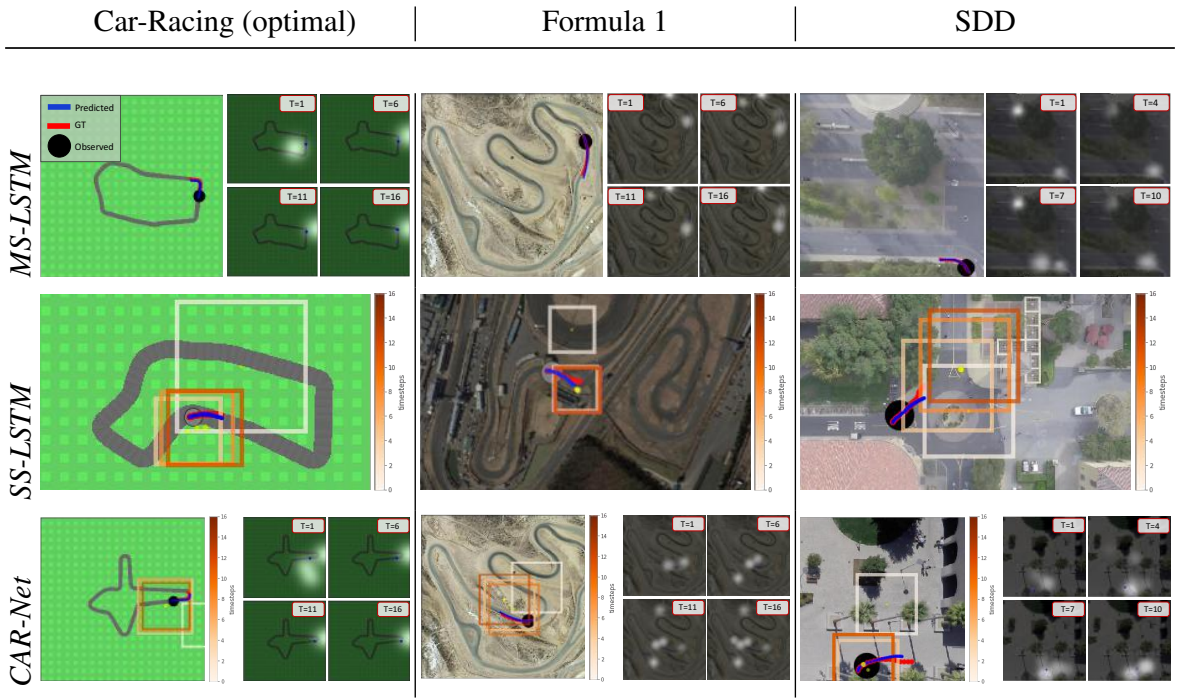

**Fig. 5.** Qualitative results of MS-LSTM, SS-LSTM, and CAR-Net (rows) predicting trajectories on Car-Racing, F1, and SDD datasets (columns). CAR-Net successfully leverages both single-source and multi-source attention mechanisms to predict future paths.

**Quantitative analysis of the impact of agent-space interactions.** To analyze CAR-Net's ability to leverage agent-space interactions, we split the test set of SDD into scenes whose geometries are complex and likely to influence the trajectories of agents (*e.g.* scenes with grass lawns, pavements, buildings), and scenes whose observable context varies little across the top-view image (*e.g.* an open field with no roads, grass, etc). We refer to these scenes as semantically *complex* and *simple*, respectively. Details about the splitting process and sample images of complex and simple scenes can be found in the supplementary material. We tested CAR-Net (that uses the scene context) and T-LSTM (that does not use any scene context) on SDD's semantically complex and simple test scenes. Our results are reported in Table 3. CAR-Net and T-LSTM achieve similar performance on *simple* scenes, where scene semantics should not typically affect the trajectories of agents. In contrast, CAR-Net achieves much better performance than T-LSTM on *complex* scenes, where scene semantics are likely to highly influence the trajectories of agents. This experiment shows CAR-Net's ability to successfully leverage scene context over T-LSTM.

| Model | Complex | Simple |
|---|---|---|
| *T-LSTM* | 31.31 | **30.48** |
| *CAR-Net* | **24.32** | 30.92 |

Table 3: Performance of T-LSTM and CAR-Net on SDD semantically complex and simple scenes. We report the Mean L2 error (ML2) in pixels space of the original image. Our method, CAR-Net, is able to effectively use the scene context to predict future trajectories.

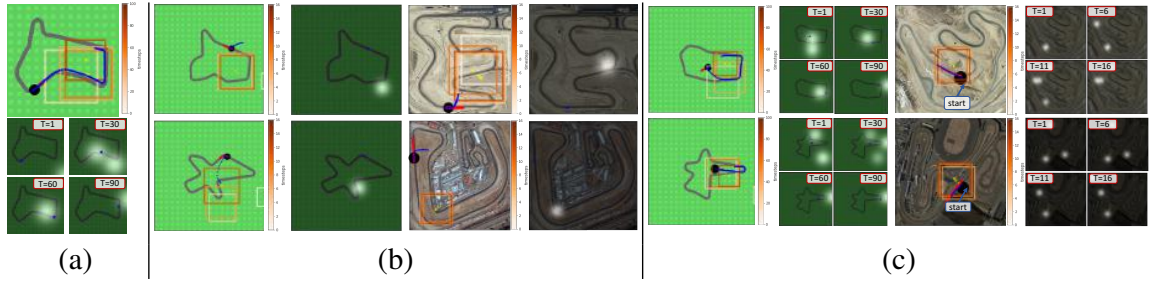

**Fig. 6.** Qualitative analysis: (a) Very long-term path prediction on the Car-Racing dataset. Predictions stay on track, showing that our model successfully uses scene context for path prediction. (b) By manually moving the attention to other parts of the image, we show that prediction heavily depends on the scene geometry. (c) When manually imposing the initial car position to be off-road, the predicted trajectory comes back on the road using the visual cues.

### 4.5 Qualitative analysis

**Visualization details.** In all figures, ground-truth and predicted trajectories are plotted in red and blue, respectively. Past positions are circled in black. We display the weight maps of the multi-source attention mechanism over time by white highlights. The single-source attention grids are also displayed over time: yellow dots represent the centers of the grids, and rectangles represent bounding boxes of the attention grid.

**Short-term predictions.** Fig. 5 shows sample trajectories predicted by our models on the datasets used in our experiments. On racing track datasets (Car-Racing and F1), we expect the region of the road close to the car to contain salient semantic elements. We observe that MS-LSTM successfully attends to the area around the car. On the mid-left and mid-center figures, we observe that the attention grid of SS-LSTM is initially off (white rectangle), before jumping to a small area close to the car, thereby identifying the relevant visual information. As shown in the bottom row, CAR-Net focuses on a narrow region of the image close to the car, using the single-source attention. It is also able to attend to further areas such as the next curve, using multi-source attention, proving its ability to leverage both attention mechanisms on racing track datasets.

On SDD, where key semantic elements are scattered, the multi-source attention mechanism successfully attends to multiple relevant visual areas (top and bottom right images). We observe that on SDD the multi-source attention attends to regions that get larger over time. This may reflect a growing prediction uncertainty. The single-source attention grid attends to areas further ahead of the agent on SDD, compared to racing track datasets (*e.g.*, the mid-right figure). It shows that attending only close to the agent would not capture all salient semantics so attention grids reach ahead.

**Very long-term trajectory prediction on Car-Racing dataset.** In this section we present qualitative results on the task of predicting future positions beyond 4.8s on the Car-Racing dataset, as a complementary result. We do not claim that our model achieves similar path prediction performance beyond 4.8s on real-world datasets. Fig. 6(a) shows CAR-Net's predictions of 100 consecutive time steps of a middle trajectory on the Car-Racing dataset. We observe that predictions remain on the road over time.

Note that the initial few positions of the agent are not helpful to predict future trajectories on very long time intervals (*e.g.* after a couple turns from the initial position). The fact that the predictions stay on the road proves that CAR-Net successfully extracts semantic understanding from the scene context in this case. We observe that both single and multi-source attention mechanisms are consistent with the predicted positions over time, as they attend to the salient parts of the scene - *e.g.*, the curve in front of the car.

**Qualitative analysis of agent-space interactions.**  We further investigate the ability of our model to leverage agent-space dependencies on racing track datasets. First, we show that road geometry has a large influence on the prediction of future positions. As shown in Fig. 6(b) left, on the Car-Racing dataset, we manually place the visual attention on an irrelevant part of the road which is oriented along the top-right direction. We observe that the model predicts positions following a similar top-right axis, while the expected trajectory without any scene information would follow a top-left direction. We observe similar behaviors in the bottom-left image of Fig. 6(b). The same experiment on the real-world F1 dataset results in similar behaviors, as shown in Fig. 6(b) right.

Second, we study whether CAR-Net is robust enough to recover from errors or perturbations by manually setting the agent's past positions outside the road. The left image in Fig. 6(c) shows the result of this experiment on the Car-Racing dataset, using the model trained on the middle trajectories. The predicted future trajectory successfully comes back on the road and remains stable afterwards, showing our model's ability to recover from strong prediction errors on Car-Racing dataset. The right image in Fig. 6(c) shows a similar experiment on the real-world F1 dataset. Since this dataset is more challenging than the Car-Racing dataset, we apply a smaller perturbation to the past trajectory of the agent, moving it slightly off the road. We observe that this perturbation does not affect the predicted trajectory which follows the road.

## 5   Conclusions

In this paper, we tackle the trajectory prediction task with CAR-Net, a deep attention-based model that processes past trajectory positions and top-view images of navigation scenes. We propose an attention mechanism that successfully leverages multiple types of visual attention. To study our model's ability to leverage dependencies between agents' behaviors and their environment, we introduce a new dataset composed of top-view images of hundreds of F1 race tracks where the vehicles' dynamics are largely governed by specific regions within the images (*e.g.*, an upcoming curve). CAR-Net outperforms previous state-of-the-art approaches on the SDD trajectory forecasting benchmark by a large margin. By visualizing the output of the attention mechanism, we showed that our model leverages relevant scene semantic features in the prediction task.

## 6   Acknowledgement

The research reported in this publication was supported by funding from the SAIL-Toyota Center for AI Research (1186781-31-UDARO), ONR (1165419-10-TDAUZ), Nvidia, and MURI (1186514-1-TBCJE).